\documentclass[lettersize,journal]{IEEEtran}
\usepackage{amsmath,amsfonts}
\usepackage{algorithmic}
\usepackage{algorithm}
\usepackage{array}
\usepackage[caption=false,font=normalsize,labelfont=sf,textfont=sf]{subfig}
\usepackage{textcomp}
\usepackage{stfloats}
\usepackage{url}
\usepackage{verbatim}
\usepackage{graphicx}
\usepackage{cite}
\usepackage{makecell}
\begin{document}

\title{BEV-Locator: An End-to-end Visual Semantic Localization Network Using Multi-View Images}

\author{Zhihuang Zhang $^{1+}$,
        Meng Xu $^{2+*}$,
        Wenqiang Zhou$^{3}$,
        Tao Peng$^{3}$, \\
        Liang Li$^{1*}$ ~\IEEEmembership{Senior Member,~IEEE,}
        Stefan Poslad$^{2}$
\thanks{+ Equal contribution. * Corresponding author.\\
1. School of Vehicle and Mobility, Tsinghua University, Beijing 100084, China. \\
2. School of Electronic Engineering and Computer Science, Queen Mary University of London, London E1 4NS, U.K.\\
3. Qcraft Inc. Beijing, 100084, China.\\
This paper was done during the internship of Zhihuang Zhang and Meng Xu in Qcraft Inc.}

}

\markboth{Journal of \LaTeX\ Class Files,~Vol.~14, No.~8, August~2021}%
{Shell \MakeLowercase{\textit{et al.}}: A Sample Article Using IEEEtran.cls for IEEE Journals}


\maketitle

\begin{abstract}
Accurate localization ability is fundamental in autonomous driving. Traditional visual localization frameworks approach the semantic map-matching problem with geometric models, which rely on complex parameter tuning and thus hinder large-scale deployment. 
In this paper, we propose BEV-Locator: an end-to-end visual semantic localization neural network using multi-view camera images. Specifically, a visual BEV (Birds-Eye-View) encoder extracts and flattens the multi-view images into BEV space. While the semantic map features are structurally embedded as map queries sequence. Then a cross-model transformer associates the BEV features and semantic map queries. The localization information of ego-car is recursively queried out by cross-attention modules. Finally, the ego pose can be inferred by decoding the transformer outputs. 
We evaluate the proposed method in large-scale nuScenes and Qcraft datasets. 
The experimental results show that the BEV-locator is capable to estimate the vehicle poses under versatile scenarios, which effectively associates the cross-model information from multi-view images and global semantic maps. 
The experiments report satisfactory accuracy with mean absolute errors of 0.052m, 0.135m and 0.251$^\circ$ in lateral, longitudinal translation and heading angle degree.



\end{abstract}






\section{Introduction}
\begin{figure}[htbp]
	\centering
		\includegraphics[width=0.95\columnwidth]{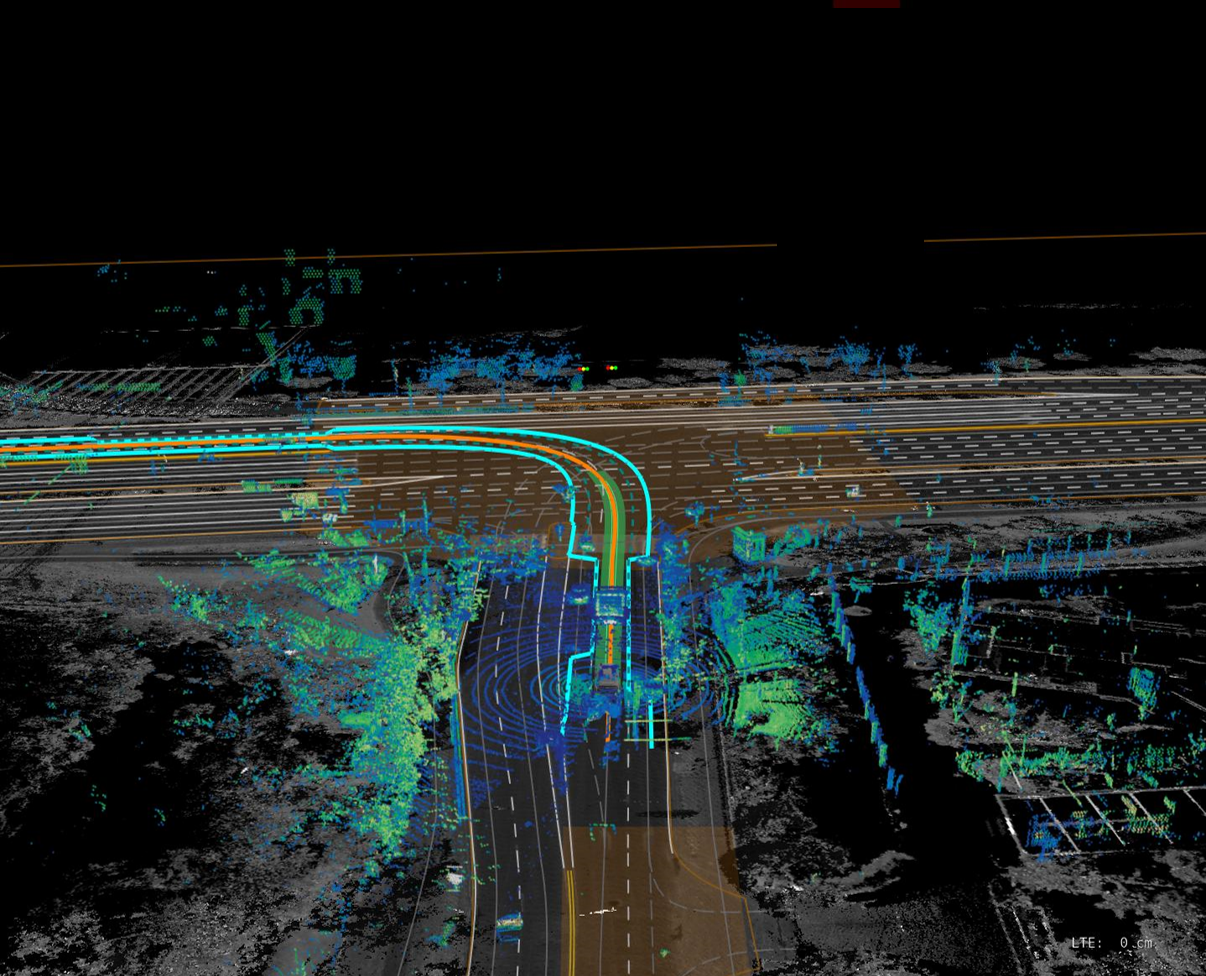}
	\caption{The perceptual range and capabilities of the intelligent vehicle are boosted from the HD map.}
	\label{fig1}
\end{figure}

\IEEEPARstart{R}{ecently} the research on intelligent driving has attracted considerable attention both in academia and industries \cite{meiring2015review, ahmed2022icran}. Accurate and robust vehicle localization is one of the crucial modules of autonomous driving and Advanced Driver Assistance Systems (ADAS) \cite{greenwood2022advanced}. As shown in Fig.\ref{fig1}, founded on the accurate pose, the perceptual range and capabilities of an intelligent vehicle can be boosted from a pre-built high-definition (HD) map. Besides, the map also provides rich fundamental prior information for vehicle navigation \cite{alkendi2021state}, planning \cite{artunedo2020motion} , and control \cite{patel2017impact}. Therefore, the task of the localization module is to estimate the precise position and orientation of the vehicle in a known scene through the sensors launched on itself. 

The problem of vehicle localization has been previously explored intensively by exploiting the features of geometry and visual appearance (e.g., scale-invariant feature transform (SIFT) features, LiDAR intensity \cite{barsan2020learning}, LiDAR point clouds \cite{yin2018locnet}, etc.), which overcomes the limitations of GPS and IMU when signal drift and blocking occur. Traditional handcrafted features (e.g., SIFT, Speeded Up Robust Features (SURF) \cite{bay2006surf}, Oriented FAST and Rotated BRIEF (ORB) \cite{rublee2011orb} features) show good performance without much variance of environment conditions, in the meanwhile, the robustness lacks under varying environments (e.g., dynamic objects, motion blur or changes in lighting). As an alternative, onboard LiDAR could acquire rich landmark information \cite{lu2019l3}. However, the high cost and expensive computation, as well as the fewer detected features and greater noises in rainy and snowy weather limit its wide application. Recent works \cite{xiao2020monocular, qin2021light, zhang2022learning, ren2021lightweight, zhao2019lidar} indicate that the semantic map with location and type information of landmarks could help to improve the robustness in localization task with a reasonable cost, Besides, the semantic description is more robust against environmental variance caused by weather change, light condition, and pavement wear, etc.

Previous works \cite{toft2018semantic, karkus2018particle, xiao2019dynamic, wu2022learning} have engaged in semantic map information based localization, which tightly links semantic and visual features. The mainstream approach implies three steps for localizing the vehicle pose: semantic feature extracting through Convolutional Neural Networks (CNN), semantic features association (e.g., RANSAC, KD-tree with semantic projection), and pose optimization \cite{cho2020robust} or filtering \cite{djuric2003particle}. While these model-based methods are fairly effective, the algorithm relies on plane assumption and hand-crafted features, which may bring projection offset and inconsistent perception ability of features from different distances and scales. To tackle these problems, previous works design complex constraints and strategies based on prior knowledge, which is effort-costly and time-consuming when processing the cross-model problem with uncertain scales.


With these challenges in mind, in this paper, we propose BEV-Locator: an end-to-end localization framework that requires little hand-engineering features extraction and parameter tuning. Instead, BEV-Locator learns to predict the optimal pose of the ego-vehicle through supervised learning. We believe this data-driven manner may significantly simplify the visual semantic localization problem. Specifically, we encode visual features by transforming surrounding images into Birds-Eye-View (BEV) feature space. In the meanwhile, the semantic map is encoded to form map queries. A transformer structure is adopted to associate map queries and BEV features. Finally the network decodes ego-pose through transformer outputs.

To the best knowledge of the authors, BEV-Locator is the first work that formulates the visual semantic localization problem as an end-to-end learning scheme. The major contributions of this research are as follows:
\begin{itemize}

\item We propose a novel end-to-end architecture for visual semantic localization from multi-view images and semantic environment, allowing accurate pose estimation of the ego-vehicle. The data-driven manner avoids geometry optimization strategy design and parameter tuning.

\item By adopting the transformer structure in cross-modal feature association, querying, and encoding-decoding. we address the key challenge of the cross-modality matching between semantic map elements and camera images.

\item We utilize the surrounding images to enhance the perceptual capabilities of the images through a unified BEV feature space. The feasibility of the visual semantic localization problem to be a subtask of BEV feature-based large model is validated.

\item Through a series of experiments on the large-scale nuScenes and QCraft datasets, we show the validity of our proposed model, which achieve the state-of-the-art performance on both dataset.
We also verify the necessity and performance of the BEV grid setting, transformer encoder strategy and positional embedding strategy by ablation study.

\end{itemize}
\section{Related Work}

\subsection{Geometric information based localization}

Geometric features have been explored to apply in large-scale visual localization with many attempts. Traditional local features (e.g., SIFT, SURF, ORB, etc.) play an important role to create 2D-3D correspondences from points in images and structure from motion (SfM) model point sets, and then the camera pose is retrieved using the matches. For instance, \cite{shotton2013scene} employs random forest to directly predict correspondences between RGB-D images and 3D scene points. \cite{li2012worldwide} proposes a bidirectional matching with SIFT features and geo-registered 3D point cloud. However, suffering from the environment with variance or repetitive conditions, the matching accuracy of local features drastically decreases especially in long-term localization. While global features (e.g., Vector of Locally Aggregated Descriptors (VLAD)-like feature \cite{jegou2010aggregating}, and DenseVLAD \cite{torii201524}) show impressive performance and robustness in terms of long-term localization, they still need training data for each scene with less scalability. 

With the development of deep learning, learnable features have recently been integrated into image matching tasks instead of hand-crafted features. The learnable descriptors (e.g., Group Invariant Feature Transform (GIFT) \cite{liu2019gift}, HardNet \cite{mishchuk2017working}, and SOSNet \cite{tian2019sosnet}, etc.) are applied after the local detection features extraction process. Furtherly, the detection and description steps are replaced into end-to-end networks, such as detect-then-describe (e.g. SuperPoint \cite{detone2018superpoint}), detect-and-describe (e.g. R2D2 \cite{revaud2019r2d2}, D2-Net \cite{dusmanu2019d2}), and describe-then-detect (e.g. DELF \cite{noh2017large}) strategies. With the filtered matches calculated from CNN feature based image matching algorithms, the localization information is then computed by RANSAC or SfM, which is computation-consuming in a multi-stage manner. Localization from the information fusion is formulated into optimization \cite{cho2020robust} or filtering problem \cite{djuric2003particle}, which relies on the plane assumption, prior knowledge, and multiple parameters in a time-consuming multi-stage manner.

Our proposed approach proposes an end-to-end framework which can be easily augmented to extract image features in changing environmental conditions (e.g., day and night, varying illumination, etc.). It also incorporates semantic information auxiliary approach, unlike some approaches that use multi-stage procedures with optimization or filtering to fuse the information, which allows the network to learn robust and accurate localization in a concise and environment-insensitive way. 

\subsection{Cross-view encoding surrounding images}
Due to the great need for facilitating the cross-view sensing ability of vehicles, many approaches have tried to encode surrounding images into the BEV feature space. In the past few years, four main types of view transformation module schemes have emerged for BEV perception. Methods (e.g., Cam2BEV \cite{reiher2020sim2real}, VectorMapNet \cite{liu2022vectormapnet}) based on Inverse Perspective Mapping (IPM) inversely map the features of the perspective space to the BEV space through the plane assumption. However, the IPM based BEV encoding method is usually used for lane detection or free space estimation because of its strict plane assumption. Another series of methods was first proposed by Lift-Splat \cite{philion2020lift}, which uses monocular depth estimation, lifts 2D image features into frustum features of each camera, and "splat" on BEV. Following work includes BEV-Seg \cite{ng2020bev}, CaDDN \cite{reading2021categorical}, FIERY \cite{hu2021fiery}, BEVDet \cite{huang2021bevdet}, and BEVDepth \cite{li2022bevdepth}, etc. Although improvements are considered in different aspects, the Lift-Splat-based method consumes a lot of video memory due to the use of additional network estimation depth and limits the size of other modules, which affects the overall performance. 

Since 2020, transformers \cite{vaswani2017attention} have become popular in computer vision, attention-based transformer shows attractiveness for modelling view transformation, where each location in the target domain has the same distance to visit any location in the source domain, and overcomes the limited local receptive field of the convolutional layer in CNN. PYVA \cite{yang2021projecting} is the first method to propose that a cross-attention decoder can be used for view transformation to lift image features to the BEV space. BEVFormer \cite{li2022bevformer} interacts spatio-temporal information through pre-defined grid-shaped BEV query, spatial cross attention (SCA) and temporal self attention. Although the data dependencies of transformers make them more expressive, they are also difficult to train. Additionally, deploying a transformer module in embedded systems with limited resources for autonomous vehicles can also be a significant challenge.

The MLP series of methods learn the mapping between perspective space and BEV space by modelling view transitions. VPN \cite{pan2020cross} stretches the 2D physical extent of the BEV into a 1-dimensional vector and then performs a full join operation on it. but it ignores strong geometric priors, each camera must learn a network with many parameters, with a certain risk of overfitting. Considering the prior knowledge, PON \cite{roddick2020predicting} proposed a semantic Bayesian occupancy grid framework, which accumulates multi-cameras information cross-image scales and timestamps. The multi-layer perceptron (MLP) method adopts a data-driven approach, and could easily employ on vehicles. Our work integrated MLP-based BEV generating mechanism \cite{li2021hdmapnet} into the visual semantic localization model to show the effectiveness of unifying the surrounding images features and connecting the BEV space, which could model 3D environments implicitly and consider the camera extrinsic explicitly, thus could be easily fused with semantic information to further improve the accuracy and robustness.
\subsection{Semantic map based localization}

Maps could provide powerful cues for tasks such as scene understanding \cite{wang2015holistic} and localization \cite{brubaker2013lost, ma2017find} in robotics using semantic labels (traffic lights, lane lines, pedestrian crossings, etc.). The pose of the vehicle could be computed by matching sensor input and a prior map, \cite{levinson2007map, levinson2010robust} represent maps as LiDAR intensities, \cite{schonberger2018semantic} constructs dense semantic maps from image segmentation and localize by matching semantic and geometric information, \cite{welzel2015improving, qu2015vehicle} follow a coarse-to-fine way, firstly the traffic sign object detection is used to retrieve the in the geometric reference database, and then the position of the vehicle is estimated through bundle adjustment. The proposal of VectorNet \cite{gao2020vectornet} provides inspiration for our network, using the vector to encode different features of the semantic map information, which enables a structured representation of semantic maps and can be adapted to the input of convolutional networks.

\subsection{Cross-model semantic and visual features association}

The semantic map-based visual localization task searches for the best matching vehicle pose by combining the current visual input and map information. The work of semantic map matching through data association and pose estimation process is beneficial to large-scale deployment with a small storage consumption, but repeatedly associating local and online semantic features brings problems such as false and missing matching. To address this problem, later works applied filtering or optimization algorithms to estimate pose, \cite{suhr2016sensor} used particle filters to update matching features, \cite{xiao2020monocular} reprojected map features and minimized line and point residuals to optimize the pose. However, with inconsistent perception ability of features at different distances and scales, these methods need prior knowledge and multiple parameters.

With the impressive continuous development of deep learning technology, transformers have investigated data association. Compared to traditional geometric feature-based methods or semantic map-based methods, learning-based localization methods combining semantic and visual features could encode useful features through neural networks without extracting flexible parameter designing and multi-stage work. 

%
We apply the transformer structure to associate visual and semantic features by cross-model querying, and further decode the vehicle pose from the query features. With the supervision of the transformer network, the model could match the semantic and visual information in an end-to-end fashion.

\section{Methods}
This section introduces the details of BEV-Locator: 
an end-to-end visual localization neural network architecture to locate the vehicle poses based on surrounding images and the semantic map. 
The visual semantic localization problem can be formulated as: Given the surrounding multi-view images $\mathcal{I}_{i=1 \hdots n}$ of the current state ($n$ indicates the number of cameras), the initial pose $[\check{x}, \check{y}, \check{\psi}]$ ($x$, $y$, $\psi$ are the 2D position and the yaw angle under local navigation coordinate system) of the ego-vehicle, and the corresponding semantic map (including the position and semantic type of boundaries, dividers, markings, poles, etc.) from online map-database, determine the optimal pose $[\hat{x}, \hat{y}, \hat{\psi}]$ of the ego-vehicle. 
Specifically, the inputs of BEV-Locator are the surrounding camera images and the semantic map that is projected to the initial pose. The output is the delta pose $[\Delta \hat{x}, \Delta \hat{y}, \Delta \hat{\psi}]$ between initial pose and the predicted pose. In other words, we obtain the optimal pose as follows:

\begin{eqnarray}
    \setlength{\abovedisplayskip}{2pt}
    \setlength{\belowdisplayskip}{2pt}
     [\hat{x}, \hat{y}, \hat{\psi}]^{T} =  
     [\check{x}, \check{y}, \check{\psi}]^{T} + 
     [\Delta \hat{x}, \Delta \hat{y}, \Delta \hat{\psi}]^{T}
\end{eqnarray}

\begin{figure*}[htbp]
	\centering
		\includegraphics[width=\textwidth]{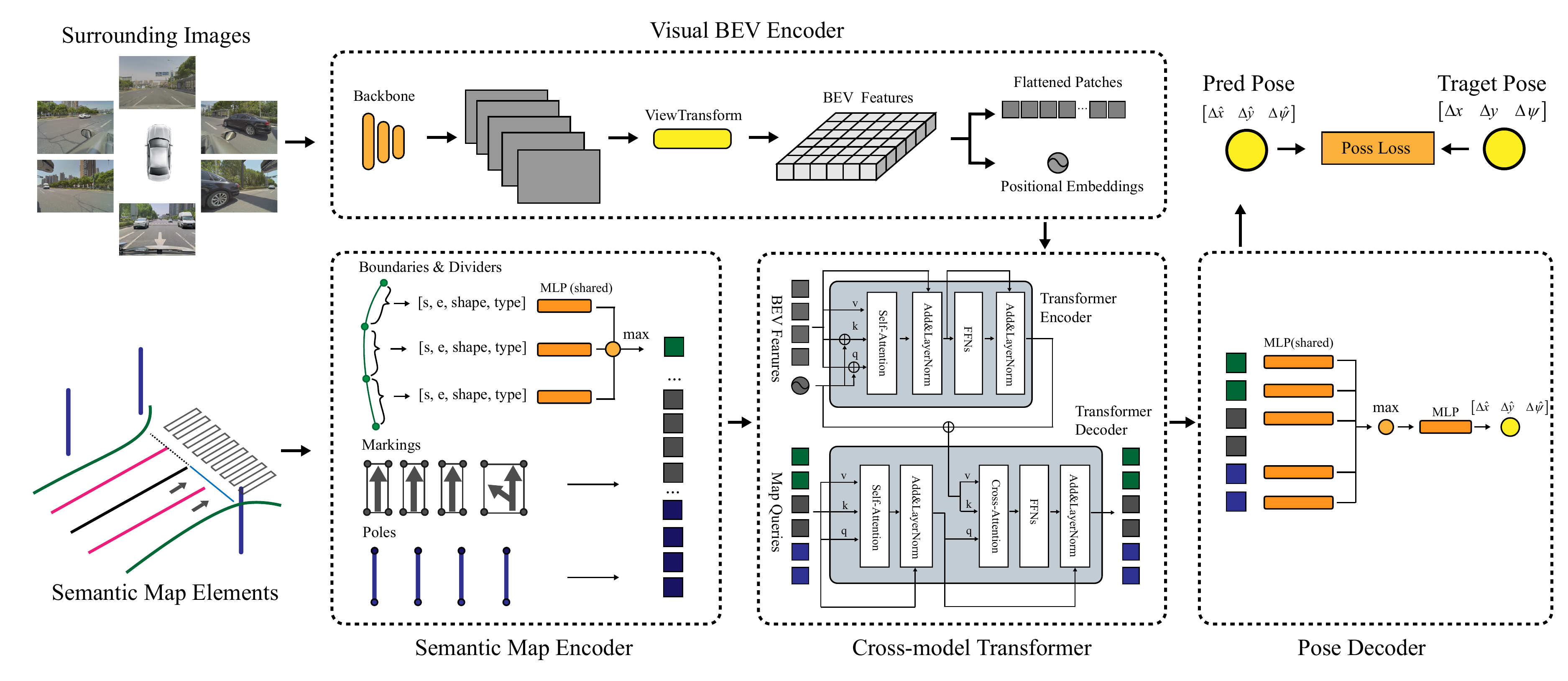}
	\caption{An overview of our proposed BEV-Locator framework, consisting of BEV Encoder (extracts features from surrounding images and projects to BEV space), Semantic Map Encoder (encodes semantic map information to structural vectors, which are also considered as map queries for transformer module), Cross-model Transformer Module (computes the attention and query ego-pose information based on map queries and BEV feature), and Pose Decoder (maps the query vectors into vehicle pose).}
	\label{fig_net_architecture}
\end{figure*}

Fig.\ref{fig_net_architecture} illustrates a modular overview of the proposed framework, consisting of a visual BEV encoder module, a semantic map encoder module, a cross-model transformer module, and a pose decoder module. The BEV feature of the surrounding images is transferred into a rasterized representation by the visual BEV encoder. The semantic map is instance-wise encoded as several compact vectors (also regarded as map queries) through the semantic map encoder. Conditioned on the BEV features and map queries, the cross-model transformer module computes the self-attention and cross-attention to query out pose information of ego-vehicle. Based on the queried-out information, the pose decoder module further infers the ego-pose where the map features have an optimal matching relationship with the corresponding images. 

\subsection{Visual BEV encoder}

The visual BEV encoder serves to extract features of images from surrounding views and project to BEV feature space, which is serially parameterized by three components, namely image feature extractor $\phi_\mathcal{I}$, view transformer $\phi_V$ and BEV feature dimensionality reduction module $\phi_R$. 

Image feature extractor $\phi_\mathcal{I}$ takes the surrounding images $\{\mathcal{I}_i \in \mathbb{R}^{C \times H_i \times W_i}\}_{i=1 \hdots n}$ from $n$ cameras as input, where $H_i$, $W_i$ and C are the dimension of height, width and channel of each image $I_i$. The feature map $\mathcal{F}_{\mathcal{I}_i}^{\phi_\mathcal{I}} \subseteq \mathbb{R}^{C_{\phi_\mathcal{I}} \times H_{\phi_\mathcal{I}} \times W_{\phi_\mathcal{I}}}$ of each image $\mathcal{I}_i$ is generated through a shared backbone, where $H_{\phi_\mathcal{I}}$, $W_{\phi_\mathcal{I}}$ and $C_{\phi_\mathcal{I}}$ represent the feature map dimension.  
\begin{eqnarray}
    \setlength{\abovedisplayskip}{2pt}
    \setlength{\belowdisplayskip}{2pt}
     \mathcal{F}_{\mathcal{I}_i}^{\phi_\mathcal{I}} = f_{\phi_\mathcal{I}}(\mathcal{I}_i)
\end{eqnarray}

Inspired by VPN \cite{pan2020cross}, we transform the extracted image features into BEV space by view transformer $\phi_V$, which contains View Relation Module (VRM) $\phi_{VRM}$ and View Fusion Module (VFM) $\phi_{VFM}$. VRM transfers the extracted image features from image coordinate to camera coordinate by MLP $\phi_{VRM}$, then with the given corresponding extrinsic matrix $\mathcal{E}_i$, the BEV feature $\mathcal{F}_{\mathcal{I}_i}^{\phi_{VFM}} $ of each image $\mathcal{I}_i$ is projected from $\mathcal{F}_{\mathcal{I}_i}^{\phi_{VRM}}$, then the BEV features $\mathcal{F}_{\mathcal{I}_i}^{\phi_{VRM}}$ of $n$ cameras are merged into the unified BEV space $\mathcal{F}_\mathcal{I}^{BEV} \subseteq \mathbb{R}^{C_{BEV} \times H_{BEV} \times W_{BEV}}$.
\begin{eqnarray}
    \setlength{\abovedisplayskip}{2pt}
    \setlength{\belowdisplayskip}{2pt}
     \mathcal{F}_{\mathcal{I}_i}^{\phi_{VRM}} = & f_{\phi_{VRM}}(\mathcal{F}_{\mathcal{I}_i}^{\phi_\mathcal{I}})  \\
     \mathcal{F}_{\mathcal{I}_i}^{\phi_{VFM}} = & f_{\phi_{VFM}}(\mathcal{F}_{\mathcal{I}_i}^{\phi_{VRM}}, \mathcal{E}_i) \\ 
     \mathcal{F}_\mathcal{I}^{BEV} = & \frac{1}{N}\sum_{i=1}^{i=N}\mathcal{F}_{\mathcal{I}_i}^{\phi_{VFM}}
\end{eqnarray}

After feature extraction, perspective transformation and external parameter transformation, we obtained dense images in the BEV space. In order to be better suitable for subsequent transformer-based network training, we then apply ResNet network $\phi_R$ to reduce the dense BEV images dimension, in which the features are reduced into a lower-resolution map $\mathcal{F}_\mathcal{I}^R$.
\begin{eqnarray}
    \setlength{\abovedisplayskip}{2pt}
    \setlength{\belowdisplayskip}{2pt}
     \mathcal{F}_{\mathcal{I}}^{R} = f_{\phi_R}(\mathcal{F}_\mathcal{I}^{BEV})
\end{eqnarray}

Inspired by the transformer structure in DETR \cite{carion2020end}, the BEV feature map $\mathcal{F}_\mathcal{I}^{R}$ are flattened into sequence as $\mathcal{F}_\mathcal{I}^t$. Besides, the model supplements the BEV features with positional embedding to preserve spatial order and enhance the perception ability. 


\begin{table}[b]
  \centering
  \resizebox{\columnwidth}{!}{
    \begin{tabular}{l}
      \hline
      \hline
      \textbf{Algorithm 1: Visual BEV Encoder}    \\
      \hline
      \textbf{Input:}\\
      \ \ \ \ $n$: Number of surrounding cameras; \\
      \ \ \ \ $\{\mathcal{I}_i\}_n$: Images from $n$ cameras; \\
      \ \ \ \ $\mathcal{E}_i$: The corresponding extrinsic matrix for each image; \\
     \textbf{Output:} \\
      \ \ \ \ $\mathcal{F}_\mathcal{I}^t$: Surrounding images flattened features in BEV space; \\
      \ \ \ \ $PE$: Positional embedding of the BEV features \\
      \ \ \ \ \textbf{1: }$\mathcal{F}_{\mathcal{I}_i}^{\phi_\mathcal{I}}\gets$Feature\_extraction$(\mathcal{I}_i)$ \\
      \ \ \ \ \textbf{2: }$\mathcal{F}_{\mathcal{I}_i}^{\phi_{VRM}}\gets$View\_Relation\_Module$(\mathcal{F}_{\mathcal{I}_i}^{\phi_\mathcal{I}})$ \\

      \ \ \ \ \textbf{3: }$\mathcal{F}_{\mathcal{I}_i}^{\phi_{VFM}}\gets$View\_Fusion\_Module$(\mathcal{F}_{\mathcal{I}_i}^{\phi_{VRM}}, \mathcal{E}_i)$ \\

      \ \ \ \ \textbf{4: }$\mathcal{F}_\mathcal{I}^{BEV}\gets$ Merge\_into\_BEV\_space $({\mathcal{I}_i}^{\phi_{VFM}}, N)$ \\
      

      \ \ \ \ \textbf{5: }$\mathcal{F}_\mathcal{I}^R\gets$Dimension\_reduction\_network $(\mathcal{F}_\mathcal{I}^{BEV})$ \\

      \ \ \ \ \textbf{6: }$(\mathcal{F}_\mathcal{I}^t, PE)\gets$Flatten\_into\_patches($\mathcal{F}_\mathcal{I}^{R}$) 
       \\
      \hline
    \end{tabular}%
    }
  \label{table_algorithm1}%
\end{table}%
\subsection{Semantic map encoder}

Semantic maps, including the elements of boundaries, dividers, road markings or poles are usually represented in the form of lines, polygons or points. However these elements lack a unified structure, therefore they could not be fed directly into neural networks. Inspired by VectorNet \cite{gao2020vectornet}, we encode the map elements from discrete points into structured vectors. 
Specifically, a semantic map $\mathcal{M}$ consists of a set of road elements.
Each element can be represented as a set of discrete points. For example, a road divider $B_i=\{v_{i}\in \mathbb{R}^2|i=1,...,N_b\}$ consists of $N_b$ points. Following the VectorNet we can denote the vector as below:

\begin{eqnarray}
    \setlength{\abovedisplayskip}{2pt}
    \setlength{\belowdisplayskip}{2pt}
    v_i = [p_i^s, p_i^e, s_i, t_i]
\end{eqnarray}
where $p_i^s$ and $p_i^e$ represent the 2D position of adjacent point inside a map element; $s_i$ stands for the shape of the map element (point, line, polygon, etc.), $t_i$ is the semantic label of the vector (road curb, road divider, pole, marking etc.). 

To form fixed-size tensors for feeding semantic map elements into network training, we design a three-dimensional structure to store semantic map element information. The first dimension size is the maximum number $D_{1}$ of map elements. The second dimension size $D_{2}$ is the maximum number of vectors (the number of discrete points in each map element). The third dimension is to represent the vector attributes. 
Following this pattern, we load the unstructured semantic map $\mathcal{M}$ into a fixed-size structured representation, where $\mathcal{M}^{t}\subseteq D_{1} \times D_{2} \times8$. We pad the blank elements with $0$ and prepare a map mask to indicate existing elements.

The semantic map encoder encodes map elements into map queries. Each node of the semantic element is first mapped to a high-dimensional space through a shared MLP. And a max pooling layer extracts the global information inside the element. In practice, the MLP and max pooling operations are repeated to increase encoder capacity. We denote the global information as a map query, which meets the concept in the transformer structure. The overall map queries is represented as $\mathcal{MQ} \subseteq \mathbb{R}^{D_{1} \times dim_{emb}}$, $dim_{emb}$ is the encoded query dimension.
\begin{table}[htbp]
  \centering
  \resizebox{\columnwidth}{!}{
    \begin{tabular}{l}
      \hline
      \hline
      \textbf{Algorithm 2: Semantic Map Encoder}    \\
      \hline
      \textbf{Input:}\\
      \ \ $\mathcal{M}$: The semantic map, which includes: \\
      \ \ \ \ $X$: Number of boundaries or dividers; \\
      \ \ \ \ $B=\{B_i\}_X$: Elements of boundaries or dividers; \\
    
      \ \ \ \ $Y$: Number of markings; \\
      \ \ \ \ $M=\{M_i\}_Y$: Elements of markings; \\

      \ \ \ \ $K$: Number of poles; \\
      \ \ \ \ $P=\{P_i\}_K$: Elements of poles; \\
     \textbf{Output:} \\
      \ \ $\mathcal{MQ}$: Map queries which is fixed-sized representation \\
      \ \ \textbf{1: }$\mathcal{M}^t \gets$Structured\_represent($\mathcal{M}$)
       \\

       \ \ \textbf{2: }$\mathcal{MQ}\gets$ Multilayer\_Perceptron\_Max\_Pooling($\mathcal{M}^t$) \\
       
      \hline
    \end{tabular}%
    }
  \label{table_algorithm2}%
\end{table}%

\subsection{Cross-model transformer module}
Our cross-model transformer module is built on the basic structure of the transformer \cite{vaswani2017attention}, which associates the map queries and BEV features to query ego-pose information. The module takes the input from the Visual BEV Encoder module and Semantic Map Encoder module and contains an encoder-decoder structure. The overall architecture of this module is depicted in Fig. \ref{fig_net_architecture}.

\paragraph{Transformer encoder} The encoder takes a flattened BEV feature patches sequence $\mathcal{F}_\mathcal{I}^t$ as input. Each encoder layer contains a multi-head self-attention module and a position-wise fully connected feed forward network (FFN), each followed by layer normalization (LN) \cite{ba2016layer} and the residual connection (RC) \cite{he2016deep}. 


\paragraph{Transformer decoder} The decoder transforms the map queries $\mathcal{MQ}$ with $D_1$ embeddings of size $dim_{emb}$. Each decoder layer consists of a multi-head self-attention module, a cross-attention module and the FFN module, each followed by LN and RC. Finally the predicted query embeddings $\mathcal{TQ}\subseteq \mathbb{R}^{D_{1} \times emb\_dim}$ are independently decoded by the FFN. Using self-attention and cross-attention mechanisms makes the model globally map the pair-wise relations between permutation-invariant map queries and BEV features, while embedding the local position information to assist the querying. Varying from the traditional transformer decoder in detection task \cite{carion2020end}, the position information of BEV features would benefit the localization task. Thus the positional embedding is also applied to the value input in the cross-attention module. This slight modification of transformer structure is major in the final accuracy, which will be discussed in the ablation study.


\begin{table}[t]
  \centering
  \resizebox{\columnwidth}{!}{
    \begin{tabular}{l}
      \hline
      \hline
      \textbf{Algorithm 3: Cross-model Transformer}    \\
      \hline
      \textbf{Input:}\\
      \ \ \ \ $\mathcal{F}_\mathcal{I}^t$: Surrounding images flattened features in BEV space; \\
      \ \ \ \ $\mathcal{MQ}$: Fixed-sized structured semantic map queries; \\
      \ \ \ \ $PE$: Positional embedding of the BEV features; \\
     \textbf{Output:} \\
     
      \ \ \ \ $\mathcal{TQ}$: Transformer guided queried feature \\
      \ \ \ \ \textbf{1: }$\mathcal{F}_{\mathcal{I},encoder}^t\gets$Transformer\_encoder$(\mathcal{F}_\mathcal{I}^t)$ \\
      \ \ \ \ \textbf{2: }$\mathcal{TQ}\gets$Transformer\_decoder$(\mathcal{MQ},\mathcal{F}_{\mathcal{I},encoder}^t, PE)$ \\

      \hline
    \end{tabular}%
    }
  \label{table_algorithm3}%
\end{table}%

\subsection{Pose decoder and pose loss function}
Conditioned on the information queried out by the transformer, the pose of the ego-vehicle can be decoded by the pose decoder. We consider each map query contains pose information or constraint offered by the corresponding map element. Therefore, the pose decoder is designed to aggregate information from each map query and predicts pose from a global perspective. We adopt a shared MLP to further encodes the map queries and a max pooling layer to aggregate global information. The max pooling layer merges the map queries into a global permutation-invariant vector. Finally, an MLP maps the global information to the offset $[\Delta \hat{x}, \Delta \hat{y}, \Delta\hat{\psi}]$ between the current estimated pose and initial pose: 
\begin{eqnarray}
    \setlength{\abovedisplayskip}{2pt}
    \setlength{\belowdisplayskip}{2pt}
    [\Delta \hat{x}, \Delta \hat{y}, \Delta\hat{\psi}] = \mathcal{PD}(\mathcal{TQ})
\end{eqnarray}

The supervision of the BEV-Locator is the ground-truth pose offset, which can be manually generated or retrieved from a more precise localization module. Given the supervision $[\Delta x, \Delta y, \Delta \psi]$ and network prediction $[\Delta \hat{x}, \Delta \hat{y}, \Delta \hat{\psi}]$, the BEV-Locator can be optimized by the Smooth L1 Loss via the following loss function:
\begin{eqnarray}
    \setlength{\abovedisplayskip}{2pt}
    \setlength{\belowdisplayskip}{2pt}
    \mathcal{L} = \alpha \times (||\Delta \hat{x}, \Delta x||_{S1} + ||\Delta \hat{y}, \Delta y||_{S1}) + ||\Delta\hat{\psi}, \Delta \psi||_{S1}
\end{eqnarray}
where $\alpha$ is the balance weight for the position loss and rotation loss. $||\cdot||_{S1}$ denotes the Smooth L1 loss.

\begin{table}[t]
  \centering
  \resizebox{\columnwidth}{!}{
    \begin{tabular}{l}
      \hline
      \hline
      \textbf{Algorithm 4: Pose Decoder}    \\
      \hline
      \textbf{Input:}\\
      \ \ \ \ $\mathcal{TQ}$: Transformer guided query feature;  
      
      \\
     \textbf{Output:} \\
     
      \ \ \ \ $[\Delta \hat{x}, \Delta \hat{y}, \Delta\hat{\psi}]$: Predicted pose offset\\
      \ \ \ \ \textbf{1: }$[\Delta \hat{x}, \Delta \hat{y}, \Delta\hat{\psi}]\gets$ Multilayer\_Perceptron\_Max\_Pooling$(\mathcal{TQ})$ \\
    \textbf{Optimization:} \\
    \ \ \ \ $\mathcal{L} = \alpha \times (||\Delta \hat{x}, \Delta x||_{S1} + ||\Delta \hat{y}, \Delta y||_{S1}) + ||\Delta\hat{\psi}, \Delta \psi||_{S1}$ \\
      \hline
    \end{tabular}%
    }
  \label{table_algorithm4}%
\end{table}%

\begin{figure}[htbp]
	\centering
            \includegraphics[width=\columnwidth]{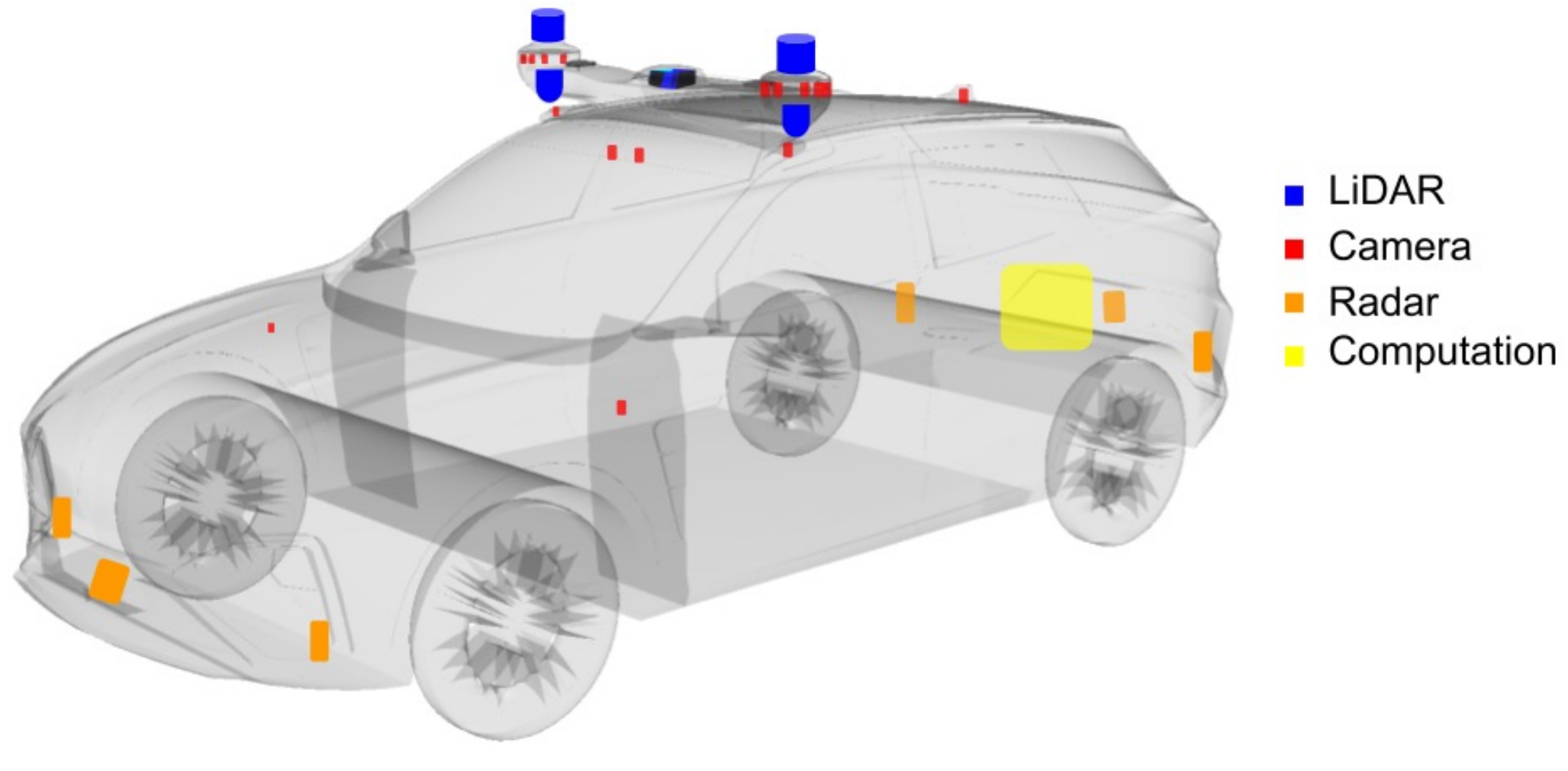}
	\caption{The experiment platform of Qcraft dataset equipped with onboard sensors and reference devices on MarvelR car.}
	\label{fig_sensor}
\end{figure}

\section{Experiments and Discussions}
\subsection{Experimental settings}
\subsubsection{Datasets}
\paragraph{nuScenes dataset} \cite{caesar2020nuscenes} is a well-known large-scale dataset for multiple autonomous driving tasks (e.g., 3D object detection task, object tracking task, etc.), which consists of 1000 scenes covering 242km collected in Boston and Singapore, each scene contains a full sensor-suite consisting of 1 LiDAR, 5 Radar, 6 cameras, IMU, and GNSS receivers, and 11 semantic layers (crosswalk, sidewalk, traffic lights, stop lines, lanes, etc.). 1.4M images were captured in RGB format with a frequency of 12Hz and a resolution of $1900 \times 900$ pixels in a diverse set of challenging locations, changing times and weather conditions (e.g., nighttime and rainy environments). The corresponding ground truth vehicle poses were calculated through a Monte Carlo Localization scheme from LiDAR and odometry information \cite{chong2013synthetic} in an offline HD LiDAR points map.
\paragraph{Qcraft dataset} is recorded by an autonomous MarvelR car \cite{marvelR} in Suzhou, China over 400km. 
Qcraft dataset was collected by 7 surrounding cameras, 5 LiDAR, 5 Radar, Fig.\ref{fig_sensor} shows the layout of the sensors. The corresponding semantic elements (lane boundary, lane divider, light pole, etc.) with positions in the global coordinate system are also supplied. The dataset includes 7 trajectories, each trajectory consists of 7 sets of images from the surrounding views. During the dataset generation, we corp and resize the images into a uniform resolution of $1920 \times 1080$ pixels. The corresponding ground truth ego poses were obtained from the trajectory of RTK and post-processing.

\subsubsection{Task and Metrics}
Visual semantic localization task takes vectorized semantic map features and surrounding images as input to estimate the vehicle localization. We measure our proposed approach with the localization metric.

Localization Metric is computed in the ego-vehicle coordinate system and measures the 3-DoF pose differences. We use the mean absolute error (MAE) and 90\% percentile values of the $[\Delta x, \Delta y, \Delta \psi]$ error to evaluate our 3-DoF localization model, which describes the localization performance of lateral, longitudinal, and yaw estimation. 

      
\subsubsection{Implementation details}
For the map encoder, the semantic map is loaded into $32 \times 128 \times 8$ tensor to represent all the semantic features (e.g., crosswalk, lane boundary, light pole,  traffic lights, cross-walks, etc.).  And the map is further encoded as $32 \times {dim_{emb}} $. We also generate the map mask to prevent blank embedding (all set as 0).  

Inspired by \cite{philion2020lift}, in the surrounding images encoding period, we use EfficientNet-B0 \cite{tan2019efficientnet} pre-trained on ImageNet \cite{russakovsky2015imagenet} as the visual backbone to extract the image features. Then, a series of MLPs are utilized to map the correspondences of the camera and ground plane. Finally the features in the bird's-eye-view are merged into dense features, which equals a physical range of $[60m \times 30m]$. To reduce the dimension of BEV raster map, ResNet18 convolution network with 4 blocks is adopted to reduce images from $[C_{BEV} \times H_{BEV} \times W_{BEV}]$ to $[dim_{emb} \times H_{BEV}/32 \times W_{BEV}/32]$.

Our proposed framework is implemented with PyTorch \cite{paszke2019pytorch} using the ADAM \cite{kingma2014adam} solver with an initial learning rate of $1e-5$ and weight\_decay of $1e-7$ on 4 NVIDIA Tesla V100 GPU using DataParallel. The SmoothL1Loss is used for calculating translation and rotation offsets, with the weight coefficient of translation error $\alpha = 0.04$. The hyperparameters are as following: cropped image size is $270 \times 480$, batch size = 8 (total batch size = 32), $max\_grad\_norm = 5.0$, and $dim_{emb} = 256$. 

In the training process, we generate random pose deviations. Specifically, we sample the random longitudinal deviation in [-2m, 2m], lateral deviation in [-1m, 1m] and yaw deviation in [-2$^{\circ}$, 2$^{\circ}$]. The semantic maps are firstly projected to the vehicle coordinate system and then translated by these deviations. The task of the network is to predict the deviation from the biased map and the surrounding images. Therefore, the deviations act as supervision, allowing the network to be trained in an end-to-end manner.


\begin{figure*}[htbp]
	\centering
		\includegraphics[width=\textwidth]{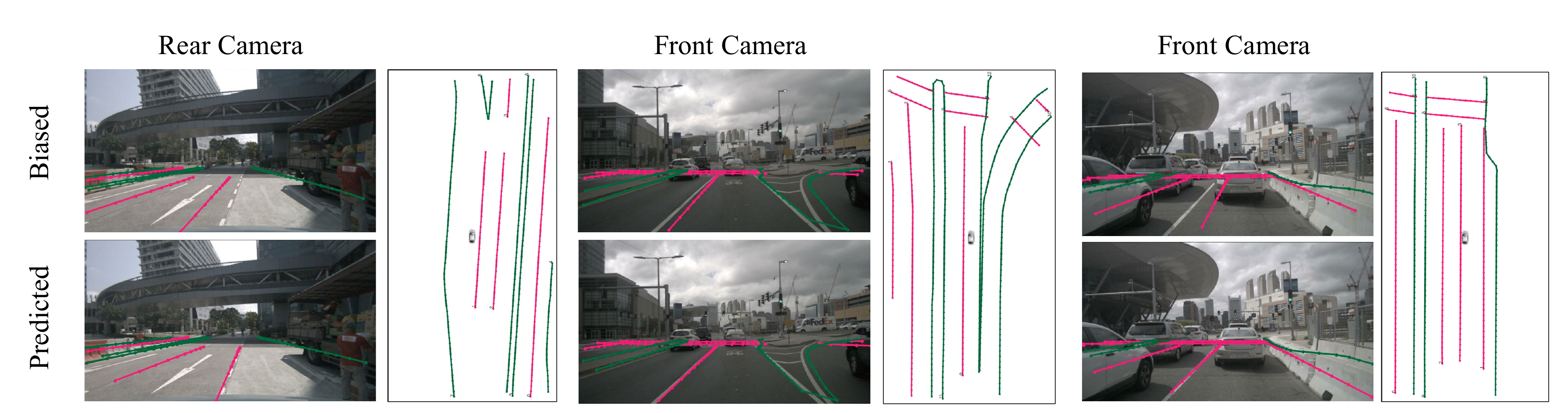}
	\caption{NuScenes results, the semantic maps are reprojected onto camera images. The lower rows show the biased poses (also known as the initial guesses) and the upper rows indicate the network predicts optimal poses where the semantic map features coincide with images.}
	\label{fig_nusence_visualization}
\end{figure*}

\begin{figure}[htbp]
	\centering
		\includegraphics[width=\columnwidth]{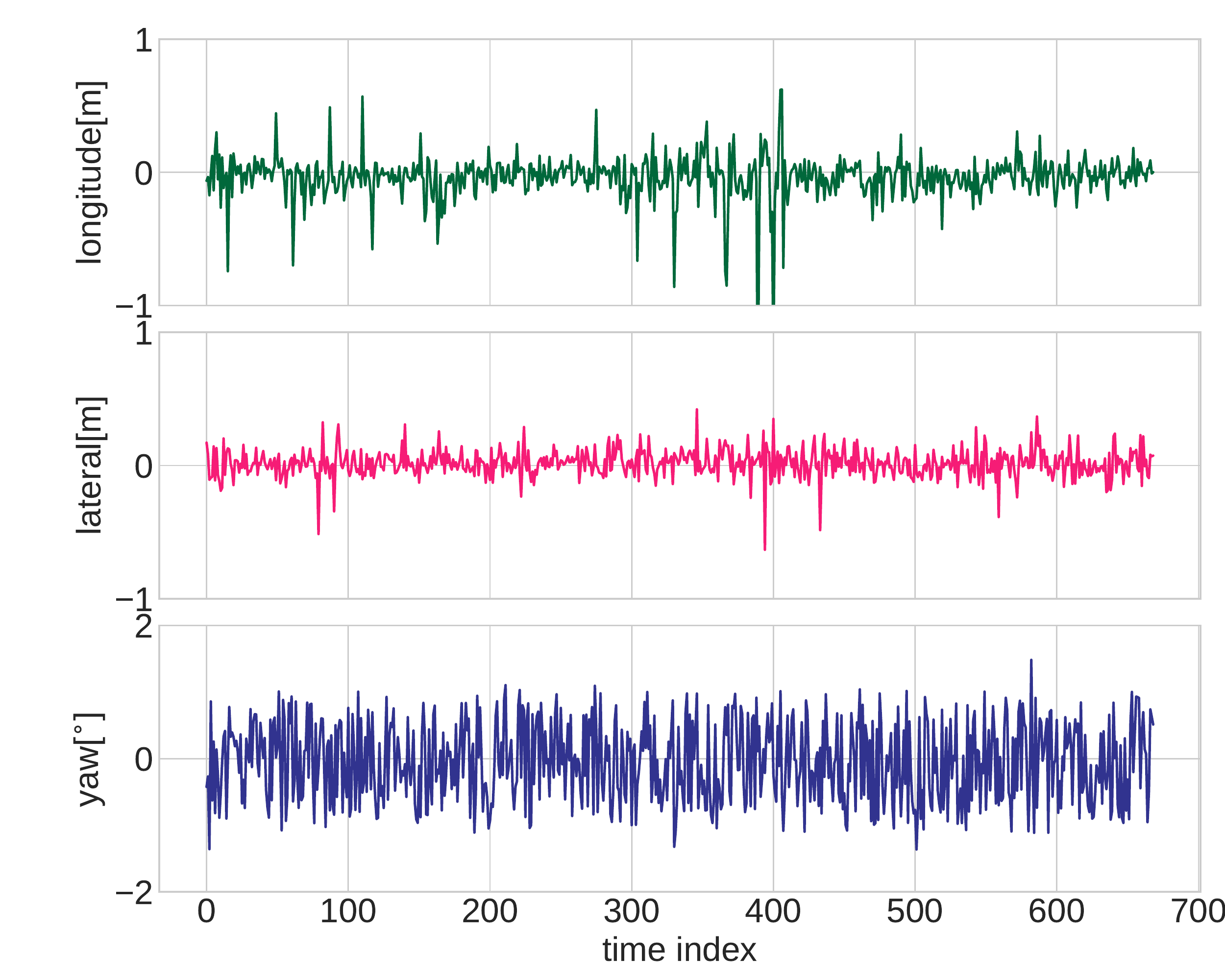}
	\caption{nuScenes dataset quantitative results.}
	\label{fig_nus_results}
\end{figure}

\subsection{nuScenes Dataset Results}

NuScenes dataset contains 700 train scenes and 150 test scenes in urban areas with images captured by 6 surrounding cameras. We conduct experiments on the nuScenes dataset to validate the effectiveness BEV-Locator (trained with 35 epochs). 

%
We extract map elements from the map interface. The element types include road boundaries, lane dividers, and pedestrian crossings. All of the 6 camera images are combined to form the BEV features. Fig.\ref{fig_nusence_visualization} visualize the localization process. Based on the provided semantic map, the initial pose, and the camera parameters, the map elements can be reprojected to the image perspective view. The upper pictures show the biased pose and the lower row pictures present the pose predicted by BEV-Locator. By comparing the upper and lower pictures, it can be observed that the map elements coincide with the elements in the camera views, which indicates the ego-vehicle present in the correct position and validates the effectiveness of BEV-Locator.

Fig.\ref{fig_nus_results} illustrate the error distribution of BEV-Locator. The error curves indicate BEV-Locator generates excellent pose accuracy. The position errors in the lateral and longitudinal directions are less than 20 cm and 60 cm.  It means the position in both lateral and longitudinal direction are well constrained by map elements in most cases. Besides, the heading directions can be predicted under 1$^{\circ}$ error. 
Through the investigation, the effectiveness of BEV-Locator in the nuScenes dataset is validated.




\begin{figure*}[htbp]
	\centering
		\includegraphics[width=0.9\textwidth]{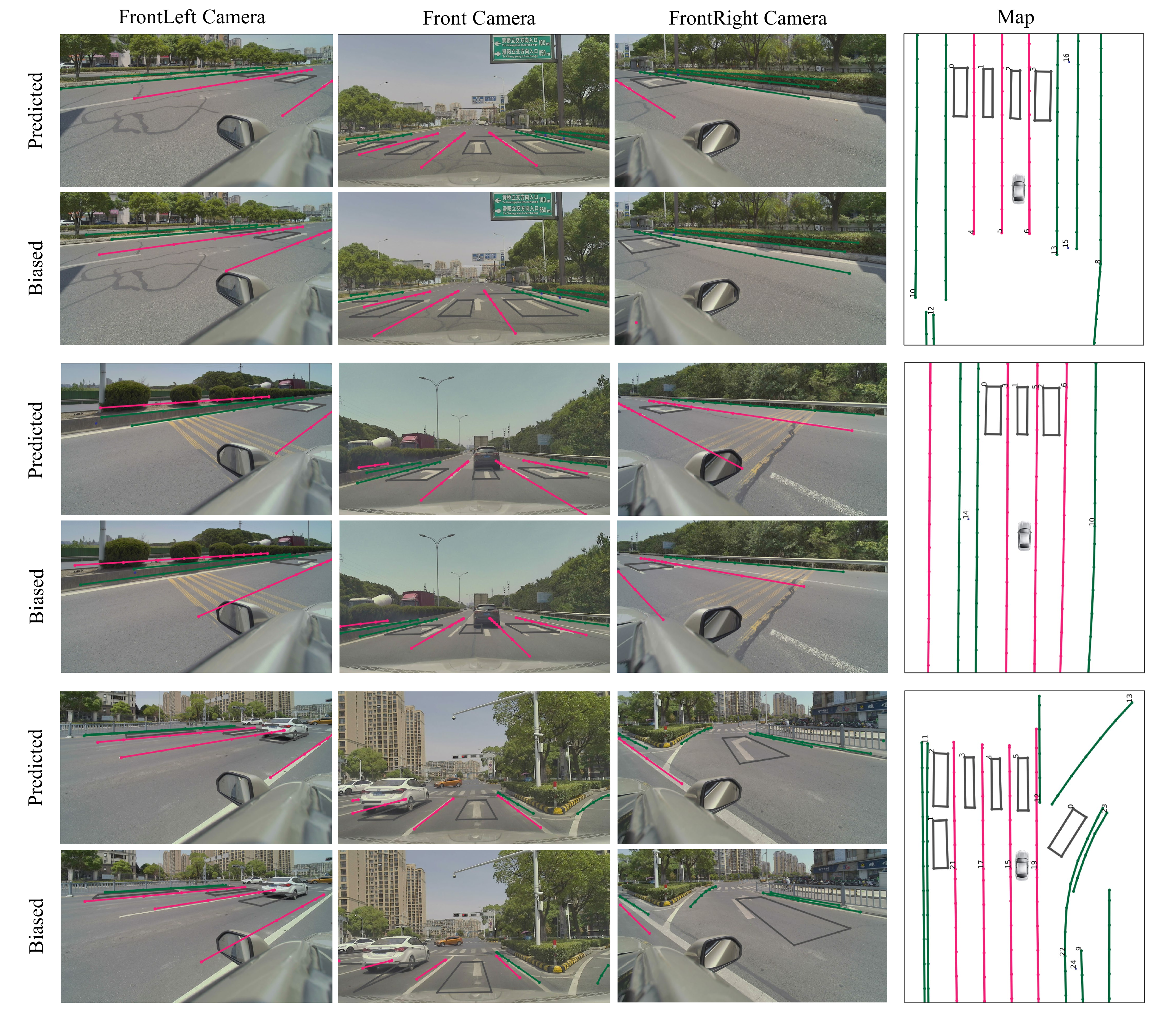}
	\caption{Qcraft results, the semantic maps are reprojected onto camera images. The lower rows show the biased poses (also known as the initial guesses) and the upper rows indicate the network predicts optimal poses where the semantic map features coincide with images.}
	\label{fig_qdataset_visualization}
\end{figure*}

\begin{figure}[htbp]
	\centering
		\includegraphics[width=\columnwidth]{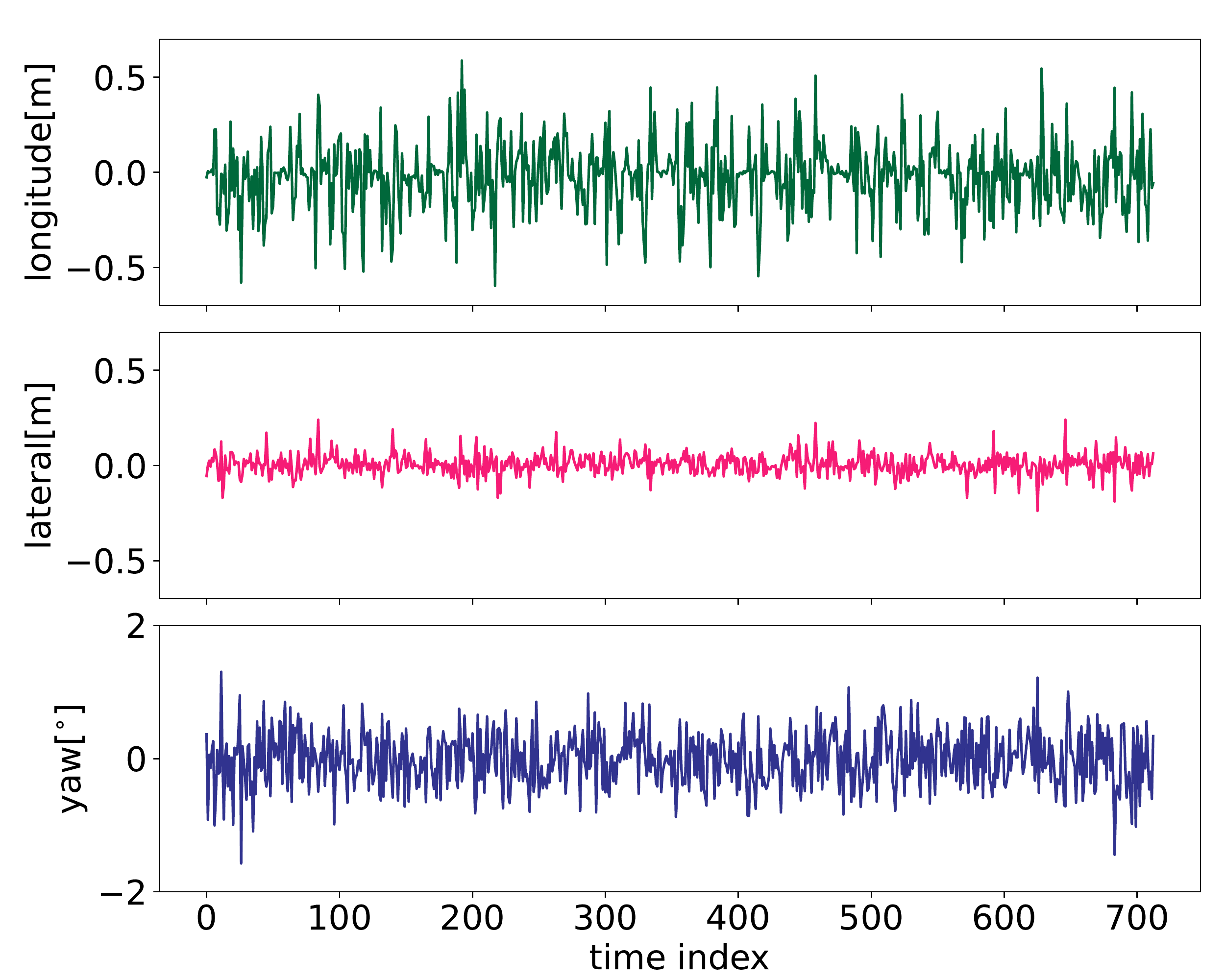}
	\caption{Qcraft dataset quantitative results.}
	\label{fig_qcraft_results}
\end{figure}

\subsection{Qcraft Dataset Results}

We further validate BEV-Locator with the Qcraft dataset, which contains urban roads and expressways with clearer lane lines and road markings. The semantic map consists of road curbs, lane dividers, road marks, and traffic light poles. For a fair comparison, 6 cameras are selected out of 7 to form the BEV features. All the training parameters are the same as those in the nuScenes dataset. 

Similarly, we show the reprojected semantic maps from three different views in Fig.\ref{fig_qdataset_visualization}. The semantic map describes the road markings with enclosed polygons and the traffic poles are presented with the contact points to the ground. It can also be concluded that the BEV-Locator successfully predicts the optimal pose of the ego-pose in scenarios of the Qcraft dataset. Combining the constraints of map elements, the position and heading of the vehicle are correctly predicted by the network. 

The error curves of a segmented trajectory are illustrated in Fig.\ref{fig_qcraft_results}. Most of the lateral and longitudinal errors are under 10 and 40 cm.  Compared with the nuScenes dataset, the BEV-Locator delivers superior accuracy in the Qcraft dataset and we ascribe this to clearer road elements and higher map quality. Next, the quantitative analysis and comparison with other methods would be discussed.

\begin{table*}[tbp]
  \centering
  \caption{Performance comparison between existing localization methods. To validate the effectiveness and performance of the proposed framework, we compare our localization results with the following methods: Choi et.al 2019 \cite{choi2019low}, Pauls et.al 2020 \cite{pauls2020monocular}, Xiao et. al 2020 \cite{xiao2020monocular}, Wang et.al 2021 \cite{wang2021visual}, Zhang et. al 2022 \cite{zhang2022learning}. }
  \resizebox{\textwidth}{!}{
    \begin{tabular}{ccccccc}
      \hline
      \hline
      Methods & Sensors & Descriptions & Dataset & \makecell[c]{Lat\\(m)}  & \makecell[c]{Lon.\\(m)} & \makecell[c]{Yaw\\($^{\circ}$)}   \\
      \hline
      
      Choi et.al 2019 \cite{choi2019low}     &\makecell[c]{Mono Camera+WSS \\+GNSS(spp)+IMU}  & Project image feature to ground plane for particle filtering  &Highway \cite{choi2019low}  &0.10 &0.25  &/ \\
      
      Pauls et.al 2020 \cite{pauls2020monocular} &Camera(mono) + WSS    &Semantic segmentation as detection front end, implicit association, pose graph optimization &Karlsruhe \cite{pauls2020monocular} &0.11 &0.90 &0.56 \\
      
      Xiao et. al 2020 \cite{xiao2020monocular} & Camera(mono) & Reprojecting map elements to image plane, optimizing the pixel distance errors & MLVHM \cite{xiao2020monocular} &  \multicolumn{2}{c}{0.29} & /\\
      
      Wang et.al 2021 \cite{wang2021visual}     &Camera(mono) + WSS + GNSS  &Novel association method and sliding window factor graph optimization &Urban road \cite{wang2021visual}&0.12 &0.43 &0.11 \\
      
      Zhang et. al 2022 \cite{zhang2022learning}                          &\makecell[c]{Camera(mono)+WSS \\+GNSS+IMU}  & Reconstructing local semantic map, matching through neural network  &Highway  \cite{zhang2022learning}& 0.054 &0.204 &/ \\
      

      \textbf{BEV-Locator}   &\makecell[c]{Multi-view Cameras} & Using single time frame images; End-to-end prediction  & nuScenes \cite{caesar2020nuscenes} & \textbf{0.076} &\textbf{0.178} & 0.510\\

      \textbf{BEV-Locator}              &\makecell[c]{Multi-view Cameras} & Using single time frame images; End-to-end prediction  & QDataset &\textbf{0.052} & \textbf{0.135} & 0.251\\
      \hline
    \end{tabular}%
    }
  \label{table_comparison}%
\end{table*}%

\subsection{Comparison with Existing Methods}
Table.\ref{table_comparison} shows the comparison of the proposed BEV-Locator against other existing localization methods. We need to affirm first that the visual localization work involves different hardware configurations, scenarios, and maps. Therefore, here we compared the overall localization accuracy of different methods.

It can be seen that BEV-Locator possesses the best position accuracy both on the nuScenes dataset and the Qcraft dataset. compared with the other approaches using multi-sensors fusion based input, our method is based on camera-only input at a single time.
In other words, our method achieves remarkable performance on the visual localization problem. Besides, since BEV-Locator can only be trained with the supervision of the pose offset, this end-to-end manner significantly simplifies the process of building a visual semantic localization system without complex strategies or parameter fine-tuning.
Moreover, since the transformer structure holds a larger learning capacity that allows for large-scale data training, the BEV-locator could be hopefully deployed to a wide range of scenarios.

We now investigate the reason that the lateral error is smaller than the longitudinal error found in the experiment results and other visual semantic localization methods.
Intuitively, the semantic elements, lane lines, road marks, and light poles provide lateral constraints simultaneously, while longitudinal position could only be constrained by light poles or road marks. The amount of longitudinal constraint elements is often less than the number of lateral constraint elements. In addition, these elements may exist at more distant distances compared with adjacent lane lines.
In summary, the longitudinal accuracy is incomparable to the lateral one. Fortunately, the downstream modules also require less accuracy for longitudinal positioning, which somewhat makes up for this problem.

\subsection{Ablation Studies} 
\label{sub_ablation}
To better understand the effectiveness of each module in our framework, we conduct the ablation study to validate through a series of comparison experiments with Qcraft dataset.

\paragraph{Effectiveness of different BEV grid sizes}

\begin{table}[htbp]
  \centering
  \caption{Effectiveness of different BEV grid sizes.}
  \resizebox{\columnwidth}{!}{
    \begin{tabular}{llll}
      \hline
      \hline
      BEV Grid Size (m) & Longitudinal (m)   & Lateral (m) & Yaw ($^{\circ}$)   \\
      0.50 & 0.200 [0.459]                & 0.072 [0.115]             & 0.269 [0.499] \\
      0.25 & 0.177 [0.400]                & 0.057 [0.118]             & 0.262 [0.427] \\
      0.15 & \textbf{0.135 [0.309]}    &\textbf{0.052 [0.107]} &\textbf{0.251 [0.395]}\\
      \hline
    \end{tabular}%
    }
  \label{table_ablation_bevsize}%
\end{table}%

To investigate the effects of different BEV grid sizes, In Table.\ref{table_ablation_bevsize} we test the impact of different BEV grid sizes on vehicle localization performance. We observe that a smaller BEV grid size contributes to higher pose accuracy. This can be explained by the fact that higher resolution allows for better encoding pose information of the map elements. However, higher resolution also brings computational burdens, posing challenges in terms of both computation time and graphics memories.

\paragraph{Effectiveness of transformer encoder}

\begin{table}[htbp]
  \centering
  \caption{Effectiveness of Transformer encoder strategy.}
  \resizebox{\columnwidth}{!}{
    \begin{tabular}{llll}
      \hline
      \hline
      Transformer Encoder & Longitudinal (m)   & Lateral (m) & Yaw ($^{\circ}$)  \\
      \hline
      Without encoder  & 0.214 [0.443]                & 0.057 [0.122]             & 0.273 [0.538] \\
      With encoder  &\textbf{0.135 [0.309]}    &\textbf{0.052 [0.107]} &\textbf{0.251 [0.395]}\\
      \hline
    \end{tabular}%
    }
  \label{table_ablation_transformer_encoder}%
\end{table}%

Table.\ref{table_ablation_transformer_encoder} exhibits the accuracy of BEV-Locator with or without the transformer encoder. Without encoder layers, the longitudinal error and lateral error drop 0.0789 m and 0.005 m, respectively.  We hypothesize that self-attention performs information interaction between BEV grids. This enables global scene awareness for road elements.

\paragraph{Effectiveness of positional embedding in transformer decoder}

\begin{table}[htbp]
  \centering
  \caption{Effectiveness of positional embedding strategy in transformer decoder.}
  \resizebox{\columnwidth}{!}{
    \begin{tabular}{llll}
      \hline
      \hline
      Transformer Decoder & Longitudinal (m)   & Lateral (m) & Yaw ($^{\circ}$) \\
      \hline
      With pos\_embedding in value     & \textbf{0.135 [0.309]}    &\textbf{0.052 [0.107]} &\textbf{0.251 [0.395]} \\
      Without pos\_embedding in value  & 1.212 [1.874]             & 0.122 [0.343]         & 0.619 [0.798]\\
      \hline
    \end{tabular}%
    }
  \label{table_ablation_transformer_decoder}%
\end{table}%

In Table.\ref{table_ablation_transformer_decoder}, we evaluate the influence of different transformer strategy in the transformer decoder module. Based on our experiments, we find that the BEV-Locator converges hardly when the conventional transformer structure is adopted, especially in the longitudinal direction.  
The problem was solved by a slight change in the transformer decoder. We add positional embedding to the value term in the cross-attention operation. Intuitively, each map query contains both semantic information and position information of the map element. 
Through the transformer, the map query is meant to query out its relative position information under BEV space. Therefore, the position information (contained in positional embedding) of each grid needs to be retrieved as a value. This small change contributes significantly to the performance of the BEV-Locator.

\subsection{Discussions}
%

To sum up, we evaluated the availability of BEV-Locator through the above experiments, from which we could conclude that our method achieves state-of-the-art performance in visual semantic localization. Conditioned on the results, we summarize the following findings:

\textbf{i)} We demonstrated that the semantic map elements can be encoded as queries. With the transformer structure, the pose information of the ego-vehicle can be queried from the BEV feature space. The effectiveness of the transformer for cross-modality matching between semantic map elements and visual images is verified.

\textbf{ii)} We formulate the visual semantic localization problem as an end-to-end learning task. The neural network requires simple supervision generated by pose offset. Simply using vehicle trajectories with raw images and the semantic map is sufficient to generate the training dataset for BEV-Locator.

\textbf{iii)} We validate the performance and accuracy of BEV-Locator on nuScense dataset and Qcraft dataset. Compared with the existing visual localization methods, BEV-Locator achieves state-of-the-art performance with only the images in a timestamp. Besides, since BEV-Locator is a data-driven method, we avoid geometry optimization strategy design and parameter tuning.

\textbf{iv)} BEV-Locator explores the feasibility of the visual semantic localization problem as a subtask of the BEV feature-based large model. Our future work aims to integrate the BEV-Locator with other perception subtasks in a large uniform BEV model. Benefiting from the BEV and transformer structure, we hypothesize that BEV-Locator has the potential to cope with large-scale scenarios.

\section{Conclusion}
We presented BEV-Locator, a new design for the visual semantic localization system based on map encoding, BEV features, and transformers for direct pose estimation of ego-vehicle. The introduced networks could efficiently encode the images and semantic maps, and further query the pose information through cross-model transformer structure. BEV-Locator is straightforward to implement following an end-to-end data-driven manner, without complex optimization strategies or complex parameter tuning. Our approach achieves state-of-the-art performance based on the nuScenes dataset and the Qcraft dataset. Our work demonstrates the effectiveness of estimating ego-pose in the BEV space. This allows visual semantic localization to be one of the subtasks of the BEV-based large model for autonomous vehicle design.
\section*{Acknowledgments}

This research was funded in part by a Ph.D. scholarship funded jointly by the China Scholarship Council and QMUL. We thank our colleagues from Qcraft who provided insight and expertise that greatly assisted the research.



\bibliographystyle{IEEEtran}
\bibliography{cas-refs}

\begin{IEEEbiography}[{\includegraphics[width=1in,height=1.25in,clip,keepaspectratio]{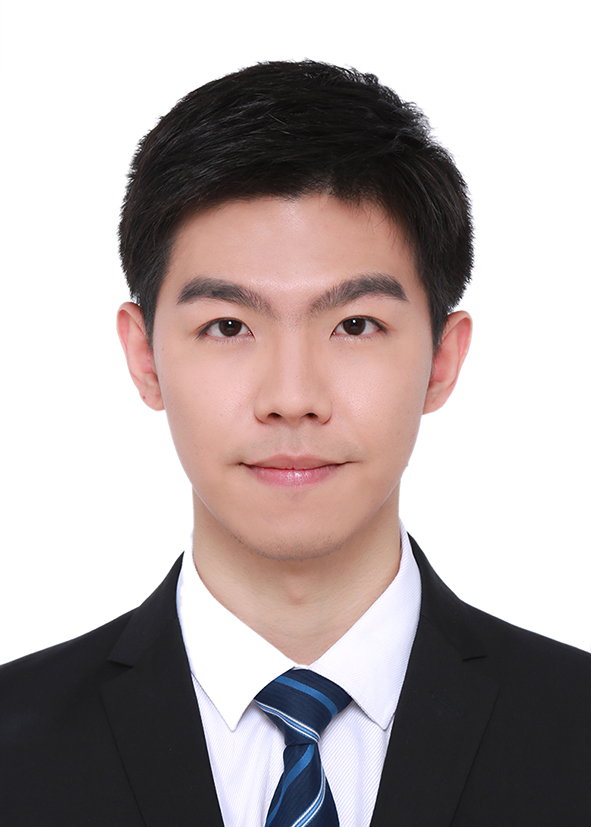}}]
  {Zhihuang Zhang} received the B.E degree in the School of Vehicle and Mobility, Tsinghua University, Beijing, in 2018. He is currently working toward the Ph.D. degree in the School of Vehicle and Mobility, Tsinghua University, Beijing. His research interests include robust and precise vehicle state estimation, multi-sensor integration perception  and high precision localization of the autonomous vehicles.
\end{IEEEbiography}

\begin{IEEEbiography}[{\includegraphics[width=1in,height=1.25in,clip,keepaspectratio]{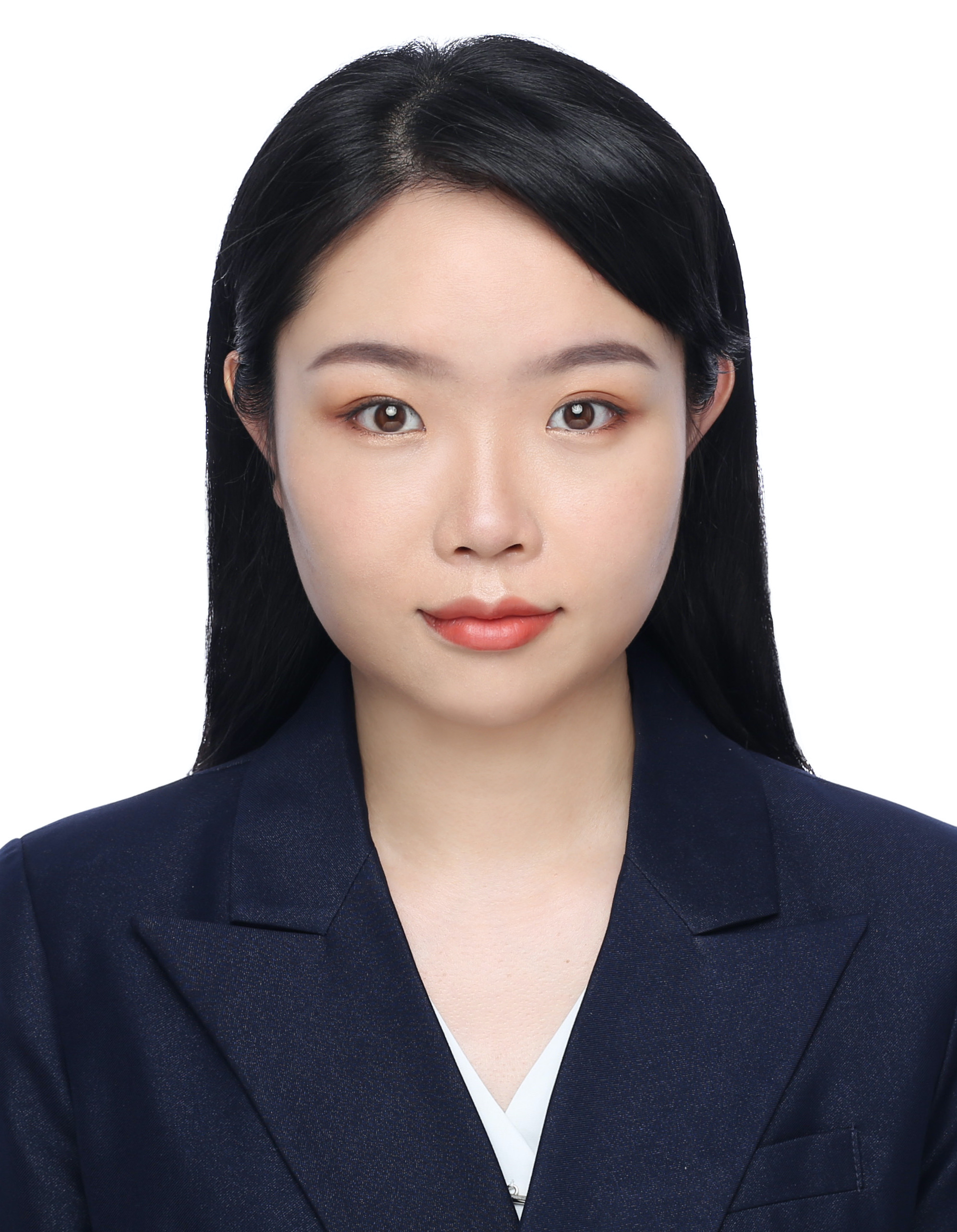}}]
  {Meng Xu} received the B.E degree in Northwestern Polytechnical University, Xi'an, in 2018. She is pursing the Ph.D. degree in the School of Electronic Engineering and Computer Science, Queen Mary University of London, London. Her research interests are spatial intelligence, machine learning and deep learning, augmented reality and urban computing.
\end{IEEEbiography}

\begin{IEEEbiography}[{\includegraphics[width=1in,height=1.25in,clip,keepaspectratio]{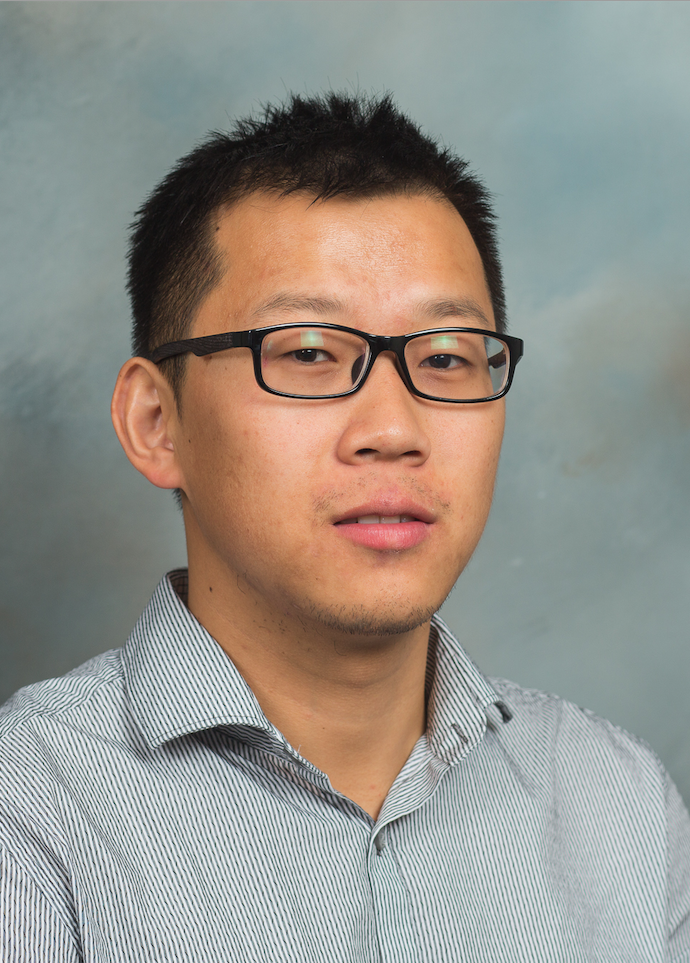}}]
  {Wenqiang Zhou} received the master's degree from Carnegie Mellon University in 2016. Since 2022, he has been a software engineer with Qcraft. His interests are vehicle localization, HD maps and robotics.
\end{IEEEbiography}

\begin{IEEEbiography}[{\includegraphics[width=1in,height=1.25in,clip,keepaspectratio]{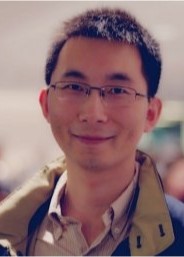}}]
 {Tao Peng} received the Ph.D. degree from Carnegie Mellon University in 2011. Since 2019, he has been an autonomous driving principal engineer with Qcraft. His interests mainly include large-scale mapping for localization, motion planning and machine learning.
\end{IEEEbiography}

\begin{IEEEbiography}[{\includegraphics[width=1in,height=1.25in,clip,keepaspectratio]{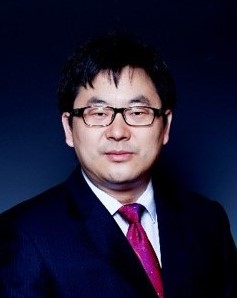}}]
  {Liang Li}received the Ph.D. degree in mechanical engineering from the  Department of Automotive Engineering, Tsinghua University, Beijing, China, in 2008. 
  
  Since 2011, he has been an Associate Professor with Tsinghua University. From November 2011 to December 2012, he was a Researcher with the Institute for Automotive Engineering, RWTH Aachen University, Aachen, Germany. His research interests mainly include vehicle dynamics and control, adaptive and nonlinear system control, and hybrid vehicle develop and control. Dr. Li was the recipient of the China Automotive Industry Science and Technology Progress Award for his achievements in the hybrid electrical bus in 2012, and the National Science Fund for Excellent Young Scholars of the People's Republic of China in 2014.
\end{IEEEbiography}

\begin{IEEEbiography}[{\includegraphics[width=1in,height=1.25in,clip,keepaspectratio]{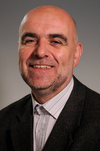}}]
  {Stefan Poslad} has a Ph.D. in medical sensing.
He is in the School of Electronic Engineering and
Computer Science, Queen Mary University of London, where he is director of the QMUL IoT Lab.
His research and teaching interests include ubiquitous computing, Internet of Things (IoT); smartenvironments, artificial intelligence and distributed systems. He has been the lead researcher for QMUL on over 15 international collaborative projects with industry, across the transport, health and environment science domains.
\end{IEEEbiography}
\end{document}